\documentclass[11pt,a4paper]{article}
\pdfoutput=1
\usepackage{arxiv}
\usepackage{times}
\usepackage{latexsym}
\usepackage{CJKutf8}
\usepackage{graphicx}
\usepackage{url}
\usepackage{amssymb}
\usepackage{mathrsfs}
\usepackage{amsmath}
\usepackage[utf8]{inputenc}
\DeclareUnicodeCharacter{FB01}{fi}

\aclfinalcopy 

\title{Evaluating Chinese-Vietnamese Low-resource Machine Translation with Back Translation}

\author{Hongzheng Li \and Heyan Huang \\
         School of Computer Science, Beijing Institute of Technology \\
         {\tt \{lihongzheng, hhy63\}@bit.edu.cn} \\}
    
\date{}

\begin{document}
\maketitle
\begin{abstract}
Back-translation (BT) has been widely used and become one of standard techniques for data augmentation in Neural Machine Translation (NMT), BT has proven to be beneficial for improving the performance of translation effectively, especially for low-resource scenarios. While most works related to back-translation mainly focus on European languages, few of them study languages in other areas around the world. In this paper, we investigate the impacts of BT on Asia language translations between the extremely low-resource Chinese-Vietnamese language pair. We evaluate and compare the effects of different sizes of synthetic data on both NMT and Statistical Machine Translation (SMT) models for Chinese-Vietnamese and Vietnamese-Chinese, with character-based and word-based settings. Some conclusions from previous works are partially confirmed and we also draw some other interesting findings and conclusions, which are beneficial to understand back-translation further.
\end{abstract}

\section{Introduction}
The great success of Neural Machine Translation (NMT) is heavily dependent on large scale parallel data. However, parallel corpora with both good quality and quantity are not always available especially for low-resource language pairs. Under the scenario where bi-texts are limited but much larger amounts of monolingual data are available, there have been extensive works to improve models with monolingual data. In which back-translation (BT) has been wildly used, and is helpful in improving the performance of translation effectively, especially for low-resource languages.

BT is a simple yet effective approach, for a translation goal from source language \textit{S} to target language \textit{T}, it first trains another intermediate system to translate the target monolingual data(\textit{T}) into the source language(\textit{S'}),  and then \textit{S'} and \textit{T} will be combined as new parallel corpus, known as the \textit{synthetic} corpus,  which will be added to the authentic parallel data to train a final system from \textit{S} to \textit{T}. BT has become one of standard technologies in the pipeline of NMT. 

BT was originally introduced to phrase-based machine translation and has been recently proposed for NMT in 2016. Since then, various works have been focusing on improving the performance of machine translation with BT, as a method of data augmentation. However, most of the works pay more attention to European low-resource languages instead of those in other areas around the world, such as Asia. We believe that some conclusions based on these languages don not necessarily suitable for all other low-resource languages. 

In this paper, we investigate and evaluate effects of back-translation for machine translation between two Asia languages: Chinese and Vietnamese. The relations between China and Vietnam have always been very close, and there have been extensive exchanges in many fields. Vietnam is also one of the countries along \textit{the Belt and Road} Initiative. Thus it is necessary to improve the quality and performance of Chinese and Vietnamese machine translation. We conduct SMT and NMT experiments with character-based and word-based settings by training models with extremely low-resource data sets, providing comparisons for Chinese-Vietnamese and Vietnamese-Chinese. To the best of our knowledge, this is the first work to comprehensively study the effects of back-translation for machine translation between the two languages. 

Our main contribution in this paper is three-fold: 

(1) we present the first comprehensive and systematic comparison of the effects of synthetic data on low-resource MT specially between Chinese and Vietnamese, and draw some new conclusions. 

(2) We evaluate synthetic data in both source and target sides for the machine translation. 

(3) We try to answer the question that "which synthetic data settings are suitable for different translation directions?" and provide some recommendation based on the experimental results.

The rest of the paper are organized as follow: section 2 briefly introduce some previous works related to this paper; section 3 describe the similarities and differences in Chinese and Vietnamese; section 4 presents the experiments and analysis in detail; Finally, the paper ends with some conclusion and future work.

\section{Related Work}
In this section, we will briefly introduce some works on back-translation and Chinese$\leftrightarrow$Vietnamese machine translation.

\subsection{Back Translation}
Back-translation was first proposed for NMT in \citet{sennrich-etal-2016-improving}, and has shown its great effectiveness in improving the performance of translation. As it is particularly useful when parallel data is scarce, BT has wide application in low-resource scenarios to leverage monolingual data. \citet{gibadullin2019survey}present a detailed survey on leveraging monolingual data in NMT.

Recently many works have aroused to understand why BT is beneficial for better NMT performance. For example, \citet{edunov2018understanding} investigate several methods to generate synthetic source sentences and their respective effects in NMT. \citet{park2017building} build the NMT model only using synthetic parallel data from both source side and target side. \citet{poncelas2018investigating} draw an empirical roadmap to observe how the amounts of BT data impact the performance of the final system, they further investigate more factors of BT data in different SMT and NMT approaches, as well as the amounts of data \citep{poncelas2019combining}. Although BT is helpful, some research also argue that the performance will decrease after the size of back-translated data reach to certain point \citep{stahlberg2018simple}.

\subsection{Chinese $\leftrightarrow$ Vietnamese Machine Translation}
There are not too many works on Chinese-Vietnamese or Vietnamese-Chinese machine translation, in which most of them use phrase-based SMT models and few involves NMT models. \citet{zhao2013vietnamese} solve Vietnamese to Chinese MT task by adopting Chinese characters as the pivot. For Chinese-Vietnamese translation, some works focus on unknown words of name entities \citep{tran2013handling,tran2014retranslating,tran2014resolving}, word segmentation\citet{tran2016word} and other challenging problems which can have negative impacts on improving the translation. \citet{tran2016character} proposes a character-based and word-based approach for Chinese-Vietnamese SMT to address the word segmentation challenge, and some other works use syntactic information to improve the performance of translation, for example, \citet{gao2019syntax} propose an effective tree-to-tree syntax-aware method for Chinese-Vietnamese MT, \citet{tran2019preordering} present word preordering approach to adjust orders in Chinese be suitable for Vietnamese first and then train SMT models with the pre-ordered data.

To the best of our knowledge, there haven't works on comparison of impacts of synthetic data on translations between Chinese and Vietnamese. 

\section{Comparison between Chinese and Vietnamese}
As the national language of Vietnam,  Vietnamese is written in Latin alphabets with additional diacritics for tones and certain letters since 20th century, and a phonetic syllable corresponds only to a single Vietnamese word. 

Vietnamese has close relations with Chinese and has been deeply impacted by Chinese characters and Chinese culture historically. Just like Japanese and Korean, Vietnamese also borrows various of characters and words from China. There are still many Sino-Vietnamese words in nowadays.

Next, we will analyze some obvious similarities and differences between the two languages with some examples.

\subsection{Similarities}
\begin{CJK*}{UTF8}{gbsn}
Syntactically, both of them are analytic languages, and have the basic SVO grammatical structures. As a result, they typically lack morphological changes, and express grammatical structures and meanings mainly by function words and word orders. 

Lexically, another similarity lies in that homonym words are very common in the two languages. That means, a Chinese character or word with the same pronunciation may have different meanings, such phenomenon is also common in Vietnamese, but happens only at monosyllabic level. For example, the three different characters (e.g., “财, 才, 材”) have different meanings (“money, talent and material” respectively), but with the same pronunciation (cái), and the three characters are all represented as the same (tài) in Vietnamese. But such phenomena may more likely to cause ambiguity and mistranslation in the process of translation, .
\end{CJK*}

\subsection{Differences}
\begin{CJK*}{UTF8}{gbsn}
First of all, the language families are different. Chinese belong to Sino-Tibetan language family, while Vietnamese belongs to Austroasiatic language family. Unlike many European languages in Indo-European languages family, these two languages seem less related in terms of language relatedness. Furthermore, they also have different written forms, so it's more difficult to mapping them into the same joint embedding representation space, and have far less shared vocabularies or BPE embedding during NMT.

\begin{figure*}[ht]
    \centering
    \includegraphics[width=16cm]{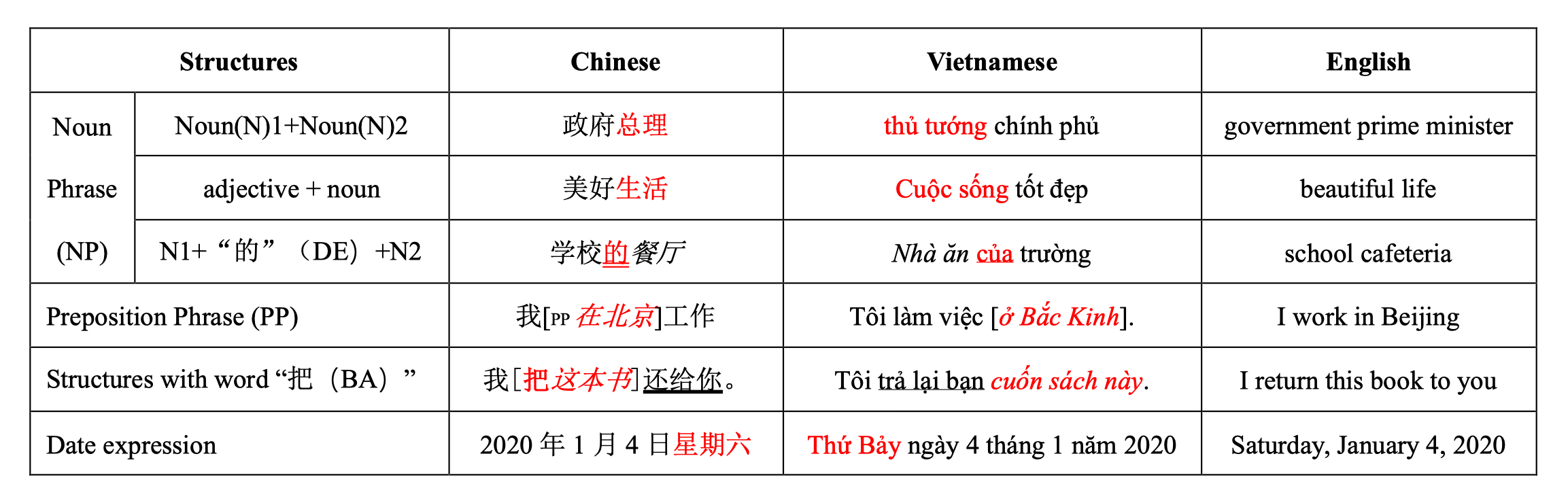}
    \caption{Some examples with different word orders in Chinese and Vietnamese}
    \label{fig:1}
\end{figure*}

Another significant difference between the two languages is the orders of many words and phrases. We list several common situations as follow where the orders are different, and corresponding examples shown in Figure \ref{fig:1}.

(1) Noun Phrase (NP). NPs in Chinese can have various grammatical structures, several of them have different orders in Vietnamese, such as: (a) "Noun(N)1+ Noun(N)2" in Chinese will be transformed to "N2 + N1" in Vietnamese; (b) "Adjective (ADJ) + N" in Chinese will also be changed to "N + ADJ" in Vietnamese; (c) NP with the function word "的（DE）". "的（DE）" is a very important function word widely used in Chinese, especially in NP. Typical structures with the word include but not limited to "N1 + DE + N2", "ADJ + DE + N" or "Pronoun + DE + N ", all of them will be reordered in Vietnamese.

(2) Position of preposition phrase (PP) in the sentence. PP usually appear after NP and before VP in Chinese sentences, i.e., sentence (S) = NP + PP + VP. But in Vietnamese, just like English, PP usually follows the VP at the end of the sentence: S = NP + VP + PP. 

(3) PP structure with the preposition “把（BA）”. This is another common but unique PP structure in Chinese, which is expressed by the special word “把（BA）”, usually following by a NP or else elements. The word has no corresponding translation in many languages including English and Vietnamese. When translating sentences with such phrase S= NP1 + \textit{把（BA）+ NP2} + VP, they
need be reordered as: NP1 + VP + NP2, and the word "把（BA）" will be deleted. 

(4) Date and week expression. The common expression format "YYYY-MM-DD-(and week name, if any)" in Chinese is expressed as “(Monday-Sunday)MM-DD-YYYY” in Vietnamese.
\end{CJK*}

The differences above, especially first three, have brought more challenges for translation between the two languages. 

Although Vietnamese is written in Latin alphabets and syllables are separated by white space, however, unlike many other Latin-alphabet based languages (e.g. English), the space cannot be used to determine word boundaries. Thus, word segmentation should be taken into consideration for machine translation and other NLP tasks. Both word-based and character-based Chinese-Vietnamese and Vietnamese-Chinese machine translation have their own advantages and drawbacks respectively. For example, as previous work discussed in \citep{tran2016character}, some entity names in Chinese, like person names (PER), must be translated as a whole into Sino-Vietnamese, which can be achieved in word-based translation, and gives better result, but word-based translation is more likely generates many unknown words. While in character-based translation, some characters in the entity may be mistranslated. 

Thus it's really necessary to investigate which translation setting is better and beneficial for MT between the two languages. This is also one of the questions we would like to investigate and answer in following section.

\section{Experiments and Analysis}

In this section, we will conduct character-based and word-based Chinese(zh)-Vietnamese(vi) and Vietnamese-Chinese translation experiments with synthetic data using SMT and NMT models. The performance of translation are compared in terms with two automatic evaluation metrics: BLEU\citep{papineni2002bleu} and METEOR\citep{banerjee2005meteor}.

\subsection{data setting}
As Chinese-Vietnamese belong to low-resource language pair, there are not many public or corpora and datasets available, some datasets used in previous works are also unavailable. As far as we know, the OPUS project\footnote{http://opus.nlpl.eu/} provides a few Chinese-Vietnamese parallel corpora such as OpenSubtitles corpora, but we found that most sentences in these corpora are too short, and the translation quality are not very good yet, such as mismatching between source and target sentences. As a result, we decide to use our own news domain data as datasets for our experiments. This also means, our results are unable to compare with those in previous works. 

The datasets contains 56,610 sentence pairs, which are all collected from multilingual news websites in Vietnam and manually processed to guarantee the quality. After shuffling, 50,000 sentences pairs are randomly selected from the corpus as training set. Then the remaining 6610 sentences are evenly divided into valid set (3305) and test set (3305) respectively. Table\ref{tab:1} shows the statistics of the datasets.

\begin{table}[htbp]
  \centering
  \caption{Statistic of datasets}
    \begin{tabular}{lccc}
    \hline
    Data  & \multicolumn{1}{p{4em}}{Sentence \newline{}pairs} & \multicolumn{1}{l}{Tokens(zh)} & \multicolumn{1}{l}{Tokens(vi)} \\
    \hline
    Training & 50,000 & 13,991 & 10,586 \\
    Dev   & 3305  & 2044  & 2101 \\
    Test  & 3305  & 1857  & 2012 \\
    \hline
    \end{tabular}%
  \label{tab:1}%
\end{table}%

In addition, we use another parallel data (about 200k sentence pairs) from various domains to train Chinese-Vietnamese SMT model with the same configuration as described in next subsection, and translate the Chinese data in the training set to generate corresponding \textit{synthetic} back-translated Vietnamese ("syn-vi" in short) data; and train Vietnamese-Chinese models for \textit{synthetic} Chinese ("syn-zh" in short). These data then will be combined with authentic Vietnamese ("auth-vi" in short) and authentic Chinese ("auth-zh" in short) data into synthetic parallel data, i.e., \{auth-zh, syn-vi\} and \{syn-zh, auth-vi\}.

Note that, the reason we don’t use the 200k corpus for datasets is because that (a), the data are combined from many domains including news, textbooks, dialogues, etc. and length of the sentences varies, and (b), the quality of the corpus is also not very good. If used as datasets, it will be difficult to determine which domain(s) of data have impacts on the final performance of translation. (c), the sentence length distribution in the news corpus is relatively uniform while have better translation quality. In addition, news domain has been one of the most used domains in the field of MT.

\subsection{models}
\begin{figure*}[ht]
    \centering
    \includegraphics[width=16cm]{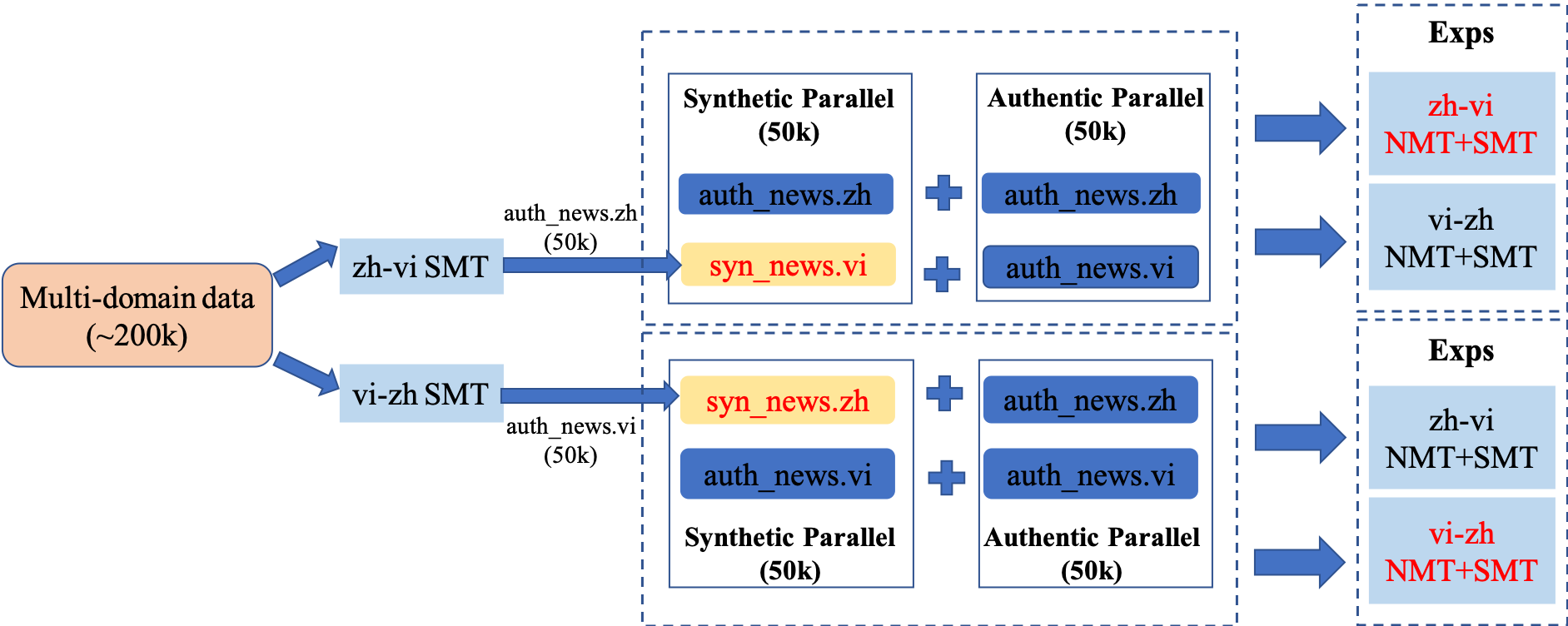}
    \caption{Pipeline of our experiments. The red parts in “Exps” indicate that the synthetic monolingual data will be used as back-translated data for the experiments.}
    \label{fig:2}
\end{figure*}

We first build word-based and character-based translation models with the authentic training set.

For SMT, we train the baseline model by using the Moses toolkit  with default settings, we use GIZA++ for word alignment and tuned with MERT \citep{och2003minimum}, and build the 5-gram language models with the KenLM toolkit\citep{heafield2011kenlm}.

For NMT, we train the baseline LSTM model using Pytorch version of OpenNMT\citep{klein2017opennmt} with default parameters, i.e., 2-layer LSTM with 500 hidden units. As the datasets are extremely small to train better models with more complicated architectures, so we decide not to use the popular Transformer\citep{vaswani2017attention}.

Inspired by \citet{poncelas2019combining}, we continue to build SMT and NMT models with increasing sizes of pseudo parallel data to evaluate the effects of back-translated data, using the same training settings with baseline models. As shown in Figure 2, the authentic 50k parallel set are considered as base data, 10k synthetic parallel data pairs are added to the base data each time to train the main models, i.e., from minimum 60k to maximum size 100k after 5 times. Note that, we keep adding synthetic data(syn-zh or syn-vi) on one side, and corresponding authentic data on other side to the new pseudo-parallel corpus each time. Under such setting, syn-zh will be used as back-translated data for zh $\rightarrow$ vi translation, and as forward-translated data for opposite direction, and so do the syn-vi data. So the data will be trained as shown in Figure\ref{fig:2}:

for Chinese-Vietnamese direction:

(1) \textit{syn-zh} + auth-zh $\rightarrow$ auth-vi + auth-vi

(2) auth-zh + auth-zh $\rightarrow$ auth-vi + \textit{syn-vi}

for Vietnamese-Chinese direction:

(3) auth-vi + auth-vi $\rightarrow$ \textit{syn-zh} + auth-zh

(4) auth-vi + \textit{syn-vi} $\rightarrow$ auth-zh + auth-zh

Besides the main models, we finally evaluate the results of SMT and NMT trained with the synthetic 50k language pair {syn-zh, syn-vi}. But their results are not the focus of our attention.

For the word-based experiment group, all the data are preprocessed with word segmentation. HanLP toolkit \footnote{https://github.com/hankcs/HanLP} is used for Chinese word segmentation, and VnCoreNLP toolkit \footnote{https://github.com/vncorenlp/VnCoreNLP} is used for segmentation of Vietnamese sentences.

\subsection{Results and Analysis}
\subsubsection{SMT Results}

The following two tables show the SMT results of two translation directions under both character-based and word-based setting.

\begin{table}[ht]
  \centering
  \caption{Results of character-based SMT}
    \begin{tabular}{p{1cm}p{1cm}{c}p{1cm}{c}p{1cm}{c}p{1cm}}
    \hline
    Directions & \multicolumn{2}{c}{zh$\rightarrow$vi} & \multicolumn{2}{c}{vi$\rightarrow$zh} \\
    \hline
    Settings & BLEU  & METEOR & BLEU  & METEOR \\
    \hline
    Baseline & 14.29 & 30.29 & 16.73 & 36.28 \\
    synthetic & 3.02 & 13.58 &  3.97 & 15.43 \\
    \hline
    \multicolumn{5}{l}{+ synthetic Chinese data} \\
    \hline
     60k  & 14.35 & 29.6  & 16.44 & 35.65 \\
     70k  & 14.81 & 30.65 & \textbf{16.70} & \textbf{36.25} \\
     80k  & 14.2  & 29.97 & 14.44 & 32.82 \\
     90k  & 14.64 & 30.38 & 16.31 & 35.79 \\
     100k & \textbf{14.95} & \textbf{30.96} & 16.3  & 35.68 \\
    \hline
    \multicolumn{5}{l}{+ synthetic Vietnamese data} \\
    \hline
     60k  & 14.46 & 29.88 & \textbf{16.88} & \textbf{36.32} \\
     70k  & 14.74 & 30.57 & 16.47 & 35.83 \\
     80k  & 14.62 & 30.36 & 14.74 & 33.11 \\
     90k  & 14.85 & 30.51 & 14.64 & 33.94 \\
     100k & \textbf{14.9} & \textbf{30.79} & 16.59 & 35.72 \\
    \hline
    \end{tabular}%
  \label{tab:2}%
\end{table}%

\begin{table}[ht]
  \centering
  \caption{Results of word-based SMT}
    \begin{tabular}{p{1cm}p{1cm}{c}p{1cm}{c}p{1cm}{c}p{1cm}}
    \hline
    Directions & \multicolumn{2}{c}{zh$\rightarrow$vi} & \multicolumn{2}{c}{vi$\rightarrow$zh} \\
    \hline
    Settings & BLEU  & METEOR & BLEU  & METEOR \\
    \hline
    Baseline & 11.71 & 26.70  & 8.58  & 23.47 \\
    synthetic  & 2.85 &  12.64  & 2.39 & 11.22 \\
    \hline
    \multicolumn{5}{l}{+ synthetic Chinese data} \\
     60k  & 11.20  & 25.79 & 7.55  & 21.63 \\
     70k  & 11.47 & 26.08 & 7.43  & 21.35 \\
     80k  & 11.34 & 25.85 & \textbf{7.60} & \textbf{21.82} \\
     90k  & 11.29 & 25.91 & 7.50  & 21.44 \\
     100k & \textbf{12.12} & \textbf{27.19} & 7.46  & 21.65 \\
     \hline
    \multicolumn{5}{l}{+ synthetic Vietnamese data} \\
    \hline
     60k  & 11.17 & 25.86 & \textbf{7.89} & \textbf{22.40} \\
     70k  & 11.10  & 25.86 & 7.34  & 21.27 \\
     80k  & 11.17 & 25.85 & 7.47  & 21.52 \\
     90k  & 11.02 & 25.73 & 7.53  & 21.57 \\
     100k & \textbf{11.20} & \textbf{25.99} & 7.53  & 21.50 \\
    \hline
    \end{tabular}%
  \label{tab:3}%
\end{table}%

As for the character-based setting in Table \ref{tab:2}, in the translation direction from Chinese to Vietnamese, adding synthetic data on both two sides can improve the performance, and all the BLEU scores are higher than authentic baseline, except the score (14.20) of the 80k synthetic Chinese data. Under each same size of synthetic data, most of scores of Vietnamese data are higher than those of Chinese, and the trend of scores is also relatively stable. However, in the opposite translation direction, strangely, synthetic data have significant negative impacts on the performance, with the adding of data size, most of BLEU scores have been declining, most of them are even lower than the baseline (16.73). When comparing the two directions, BLEU scores of vi$\rightarrow$zh are generally about2 points higher than those of zh$\rightarrow$vi.

As for the word-based setting in Table \ref{tab:3}, overall, adding either Chinese or Vietnamese data hardly enhances the translation quality regardless of the translation directions, as most of the results are just below the baseline. As shown of the results in bottom left corner of table 3, the results of 60k and 100k Vietnamese data are almost the same, even though the data size varies by half a million. On the contrary, as the data sizes increase, the scores have been fluctuating and showing downward trends. Finally, different with character-based results, the scores of vi-zh here are much lower zh-vi ones.

From above two tables, we can see that synthetic data have opposite effects on the two translation directions. More specific, both the monolingual synthetic data from two sides have positive impacts when translating Chinese to Vietnamese, furthermore, the positive effects are more pronounced in character-based translations. However, they do have negative impacts on Vietnamese to Chinese translation, especially the back-translated Vietnamese data.

When comparing the same translation directions between character-based and word-based settings, it is clearly that former settings perform much better than latter in both zh$\rightarrow$vi and vi$\rightarrow$zh directions, especially from Vietnamese to Chinese, indicating that character-based settings maybe more beneficial in SMT. Scores of character-based are about 3 points higher than those of word-based zh$\rightarrow$vi, and scores of character-based vi$\rightarrow$zh are even almost 7-9 points higher than word-based vi$\rightarrow$zh. 

\subsubsection{NMT Results}
Next, we’ll take an inside look at the NMT performance, as shown in table 4 and table 5.

\begin{table}[htbp]
  \centering
  \caption{Results of character-based NMT}
     \begin{tabular}{p{1cm}p{1cm}{c}p{1cm}{c}p{1cm}{c}p{1cm}}
    \hline
    Directions & \multicolumn{2}{c}{zh$\rightarrow$vi} & \multicolumn{2}{c}{vi$\rightarrow$zh} \\
    \hline
    Settings & BLEU  & METEOR & BLEU  & METEOR \\
    \hline
    Baseline & 11.68 & 25.32 & 13.39 & 28.57 \\
    \hline
    synthetic & 2.94  & 13.43 & 3.05  & 13.03 \\
    \hline
    \multicolumn{5}{l}{+ syn-zh} \\
    \hline
     60k  & 11.43 & 23.41 & 13.32 & 28.88 \\
     70k  & 11.89 & 24.37 & 13.67 & 29.57 \\
     80k  & \textbf{12.34} & \textbf{24.96} & 14.28 & 30.42 \\
     90k  & 12.18 & 24.78 & \textbf{14.32} & 30.65 \\
     100k & 11.84 & 24.21 & 14.31 & \textbf{30.74} \\
    \hline
    \multicolumn{5}{l}{+ syn-vi} \\
    \hline
     60k  & \textbf{11.23} & \textbf{23.77} & 13.29 & 28.48 \\
     70k  & 7.66  & 18.73 & 13.29 & 28.64 \\
     80k  & 6.49  & 17.71 & 12.68 & 27.90 \\
     90k  & 5.50  & 16.72 & 13.14 & 28.48 \\
     100k & 4.92  & 16.06 & 12.83 & \textbf{27.95} \\
    \hline
    \end{tabular}%
  \label{tab:4}%
\end{table}%

\begin{table}[htbp]
  \centering
  \caption{Results of word-based NMT}
     \begin{tabular}{p{1cm}p{1cm}{c}p{1cm}{c}p{1cm}{c}p{1cm}}
    \hline
    Directions & \multicolumn{2}{c}{zh$\rightarrow$vi} & \multicolumn{2}{c}{vi$\rightarrow$zh} \\
    \hline
    Settings & BLEU  & METEOR & BLEU  & METEOR \\
    \hline
    Baseline & 9.65  & 22.77 & 7.44  & 20.28 \\
    \hline
    Synthetic & 7.21  & 19.23 & 5.91  & 18.19 \\
    \hline
    \multicolumn{5}{l}{+ syn-zh} \\
    \hline
     60k  & 9.51  & 22.50  & 6.99  & 19.65 \\
     70k  & 9.51  & 22.58 & \textbf{7.92} & 21.07 \\
     80k  & 9.67  & 22.79 & 7.67  & 20.94 \\
     90k  & 9.89  & 22.97 & 7.69  & 20.86 \\
     100k & \textbf{9.90} & \textbf{23.00} & 7.97  & \textbf{21.50} \\
    \hline
    \multicolumn{5}{l}{+ syn-vi} \\
    \hline
     60k  & 9.75  & 23.07 & 7.14  & 19.93 \\
     70k  & 10.27 & 23.89 & 7.72  & 20.80 \\
     80k  & \textbf{10.47} & 24.29 & \textbf{7.90} & 20.87 \\
     90k  & 10.19 & 24.09 & 7.66  & 20.91 \\
     100k & 10.27 & \textbf{24.33} & 7.89  & \textbf{21.17} \\
    \hline
    \end{tabular}%
  \label{tab:5}%
\end{table}%

First, Table \ref{tab:4} shows that adding increasing sizes of back-translated synthetic Chinese data to Chinese-Vietnamese translation has a positive impact on the performance, as four of five data sizes are higher than baseline with authentic data. But we observe that the best scores of both BLEU and METEOR appear in the middle data size (80k) instead of the largest one, which is quite interesting and beyond our expectation. From 60k to 80k data, the scores increase correspondingly, but decrease since then. The similar conclusion is also proved in previous works\citep{stahlberg2018simple}. This result seems to indicate that more synthetic data is not always better, especially for some far-distance languages. When it comes to other translation direction, the positive effects of synthetic Chinese data are obvious and the increasing data sizes continue to improve the translation performance, until reaching the best status under the largest data size, about 1 point higher than baseline. 

On the other hand, effects of synthetic Vietnamese data are quite different with Chinese, it’s clearly shown that, no matter in which translation directions, the increase of Vietnamese data always leads to the decrease of scores of BLEU and METEOR. The larger the data size, the worse the translation performance, especially in the Chinese-Vietnamese direction, the BLEU scores even decrease sharply from 11 (60k data) to below 5(100k data).  

In the word-based NMT in Table \ref{tab:5}, the results, once again, prove that synthetic Chinese data are beneficial to improving the translation. Similar to the Chinese data in character-based Chinese to Vietnamese NMT, the best BLEU scores of two directions with synthetic Vietnamese data are in the middle size setting (80k), and the scores also increase but then decrease, while the best scores of METEOR are in the largest data group. The results here show that synthetic Vietnamese data have certain positive impacts on achieving improvements of the translation performance compared with baseline under word-based setting.

When comparing the same translation directions between character-based and word-based NMT, except the word-based Chinese-Vietnamese with synthetic Vietnamese data outperform the character-based ones, while all remaining character-based results are better than word-based ones, 2 points higher in Chinese-Vietnamese and about 6 points in other direction.

\subsubsection{Comparison between SMT and NMT}

We would like to discuss some comparison between SMT and NMT. Many previous works have argued that SMT results tend to outperform NMT results under low-resource data scenarios. As the datasets for our experiments are too small to train very good NMT models, further, unlike many European languages, relatedness between Chinese and Vietnamese are relatively distant, because they have different writing symbols. Intuitively, SMT results should much better than NMT, which have been proven here. 

However, we're excited to find that in word-based Vietnamese to Chinese translations, most NMT results surprisingly outperform SMT. As the comparison shown in Figure 3, eight out of ten results of NMT significantly outperform SMT in the two test groups with synthetic data.

As \citet{sennrich2019revisiting} discussed in their experiments, low-resource NMT is very sensitive to hyperparameters, if tuning well, training competitive NMT systems is possible to surpass SMT systems. But the NMT systems in our experiments are trained without any tuning, even without the popular BPE operations \citep{sennrich2015neural}. 

\begin{figure}[ht]
    \centering
    \includegraphics[width=7cm]{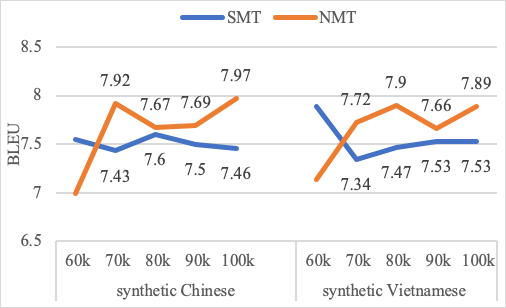}
    \caption{Comparison between word-based Vietnamese-Chinese SMT and NMT}
    \label{fig:3}
\end{figure}

We further analyze the outputs of word-based Vietnamese to Chinese SMT and NMT, and find that NMT translations seems to be more advantageous in two aspects: (a) There are less untranslated words of source language, including entity names and other words, (b) word orders and syntactic structures are more readable and suitable.

\begin{figure*}[ht]
    \centering
    \includegraphics[width=16cm]{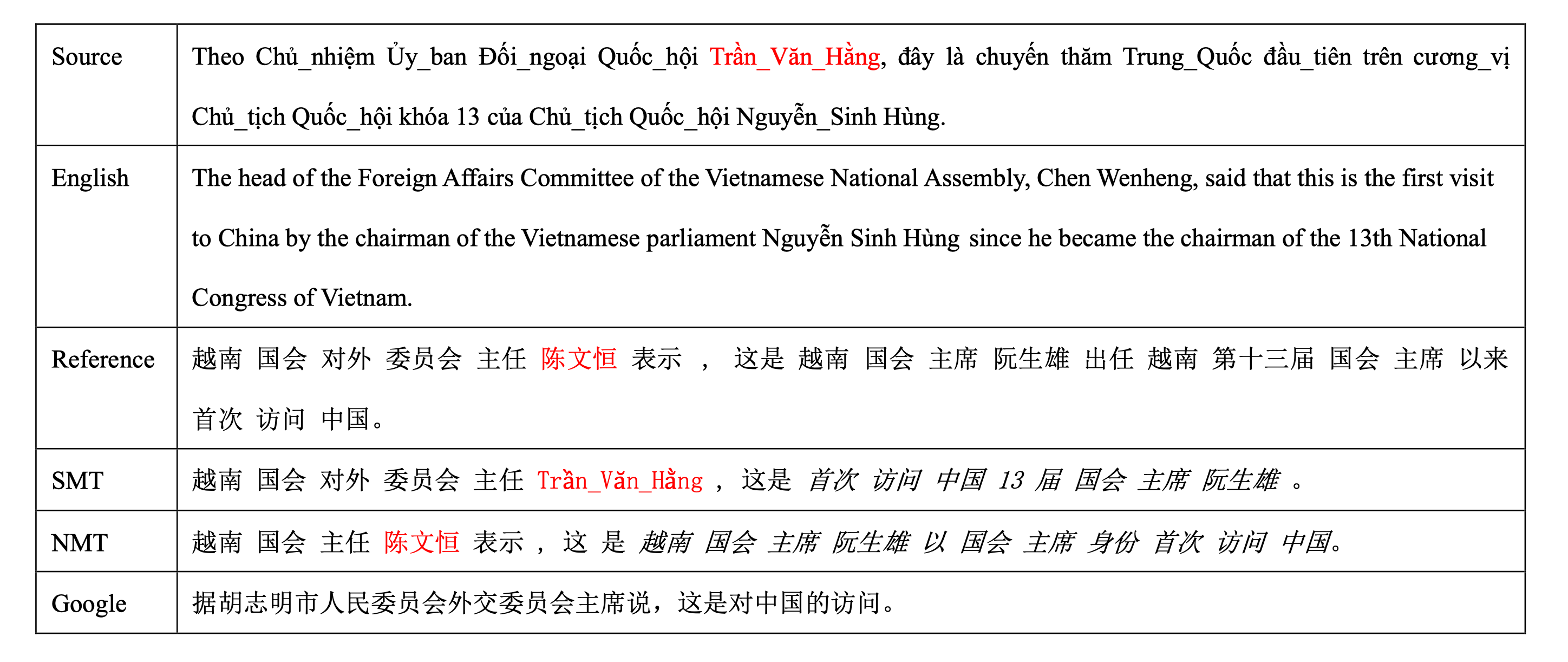}
    \caption{Translation of an example sentence in SMT and NMT}
    \label{fig:4}
\end{figure*}

Figure \ref{fig:4} shows an example sentence in Vietnamese and corresponding translations of word-based SMT, NMT, which can clearly explain why the performance of NMT in the experiments are better. In the example, the person name (in red color) is not translated by SMT, but is correctly translated in NMT; on the other hand, The syntactic structure of the italicized sentence ("The chairman visits China.") in SMT is mistranslated as "visits China the chairman", while its translation in NMT is very reasonable. We also put the source example into the Google Translate, we can see the length of the outputs is much shorter, because it suffers under-translation problem, in other words, many (important) words are not translated at all. 

Finally, based on the preliminary experimental results and discussion above, we can draw some conclusions as follow:

(1) \textbf{Adding synthetic Chinese data has positive impacts on improving the performance of two translation directions under both character-based and word-based setting.} Especially the back-translated data added to Chinese-Vietnamese translation. But on the other hand, more synthetic data is not always better. Because although performance of most of them are better than baseline, some best scores appear in middle data size instead of the largest ones.

(2) \textbf{On the whole, the impacts of synthetic Vietnamese data on the performance of translation is more prominent than positive impacts.} Back-translated data have obvious negative effects on the performance of Vietnamese-Chinese in most SMT and NMT cases, performance decrease or hardly improve with the increasing data size. The forward-translated data also severely degrade translation performance of character-based Chinese-Vietnamese NMT, while they only have certain positive impacts on character-based Chinese-Vietnamese SMT as well as in word-based Chinese-Vietnamese NMT.

(3) \textbf{We recommend character-based settings for both Chinese-Vietnamese and Vietnamese-Chinese MT.} As discussed above, from the perspective of character/word settings, under the character-based setting, both SMT and NMT for Vietnamese-Chinese outperform the other direction('$\triangle$' in Table \ref{tab:6}), while under word-based setting, the latter is better than the former; from the perspective of translation direction, scores of character-based SMT and NMT all higher than those of word-based ones in the both two directions('$\surd$' in Table \ref{tab:6} ). Thus, we believe that character-based settings are more suitable.

\begin{table}[htbp]
  \centering
  \caption{Translation settings and translation directions}
    \begin{tabular}{p{2.5cm}p{1.5cm}p{1.5cm}}
    \hline
    \multicolumn{1}{r}{} & \multicolumn{1}{p{2.835em}}{zh$\rightarrow$vi} & \multicolumn{1}{p{2.835em}}{vi$\rightarrow$zh} \\
    \hline
    character-based & $\surd$   & \multicolumn{1}{p{5em}}{$\surd$, $\triangle$ } \\
    word-based & $\triangle$     &  \\
    \hline
    \end{tabular}%
  \label{tab:6}%
\end{table}%

\section{Conclusion and Future Work}
In this paper, we present preliminary empirical investigation about the effects of different sizes of synthetic monolingual data for Asia low-resource machine translation. We discuss some similarities and differences between the Chinese and Vietnamese, then conduct comparison of SMT and NMT experiments of Chinese-Vietnamese and Chinese-Vietnamese, with both character-based and word-based setting and quite limited data, evaluating the performance of models trained with increasing size of synthetic back-translated data.

The popular conclusion from some previous works that back-translation is especially helpful for low-resource MT scenarios is partially confirmed in our experimental results, but we find that the performance will not always improve as the amounts of data increase, on the contrary, the performance can even decrease. Furthermore, the actual effects of back translation may depend on languages, data sizes and translation directions, as well as some other factors. Some interesting ﬁndings and conclusions in this paper provide good clues and directions for future work.
In the future, we would like to expand the dataset size to conduct deeper analysis further, such as the grammatical changes with synthetic data, and we will incorporate back-translation with other approaches such as transfer learning to improve translation performance of between Chinese and Vietnamese.

\section*{Acknowledgment}
We would like to thank Dr.Huu-Anh Tran for providing the Chinese and Vietnamese parallel data for this work, and we thank Dr. Alberto Poncelas, from the School of Computing, DCU, for the discussion and suggestions.

\bibliography{arxiv}
\bibliographystyle{acl_natbib}

\end{document}